\documentclass[10pt,twocolumn,letterpaper]{article}
\usepackage{btas}
\usepackage{times}
\usepackage{epsfig}
\usepackage{graphicx}
\usepackage{amsmath}
\usepackage{amssymb}
\usepackage{makecell}
\newcommand{\RNum}[1]{\uppercase\expandafter{\romannumeral #1\relax}}

\btasfinalcopy 

\ifbtasfinal\fi

\begin{document}
	
\title{Polarimetric Thermal to Visible Face Verification\\ via Attribute Preserved Synthesis}

\author{Xing Di$^1$,  \qquad He Zhang$^2$, \qquad Vishal M. Patel$^1$\\
	$^1$Johns Hopkins University, 3400 N. Charles St, Baltimore, MD 21218, USA\\
$^2$Rutgers University, 94 Brett Rd, Piscataway Township, NJ 08854, USA\\
{\tt\small xdi1@jhu.edu, he.zhang92@rutgers.edu, vpatel36@jhu.edu}
}

\maketitle
\thispagestyle{empty}

\begin{abstract}
Thermal to visible face verification is a challenging problem due to the large domain discrepancy between the modalities.  Existing approaches either attempt to synthesize visible faces from thermal faces or extract robust features from these modalities for cross-modal matching.  In this paper, we take a different approach in which we make use of the attributes extracted from the visible image to synthesize the attribute-preserved visible image from the input thermal image for cross-modal matching.  A pre-trained VGG-Face network is used to extract the attributes from the visible image. Then, a novel Attribute Preserved Generative Adversarial Network (AP-GAN) is proposed to synthesize the visible image from the thermal image guided by the extracted attributes.  Finally, a deep network is used to extract features from the synthesized  image and the input visible image for verification.   Extensive experiments on the ARL Polarimetric face dataset show that the proposed method achieves significant improvements over the state-of-the-art methods. 
\end{abstract}

\section{Introduction} \label{introduction}
Face Recognition (FR) is one of the most widely studied problems in computer vision and biometrics research communities due to its applications in authentication, surveillance and security.  Various methods have been developed over the last two decades that specifically attempt to address the challenges such as aging, occlusion, disguise, variations in pose, expression and illumination.  In particular, convolutional neural network (CNN) based FR methods have gained a lot of traction in recent years \cite{FR_SPM_2018, HyperFace}.  Deep CNN-based methods \cite{parkhi2015deep,schroff2015facenet,wen2016discriminative,JC_WACV2016,peng2016recurrent,peng2018red} have achieved impressive performances on the current FR benchmarks.

\begin{figure}[t]
	\centering
	\includegraphics[width=0.85\linewidth]{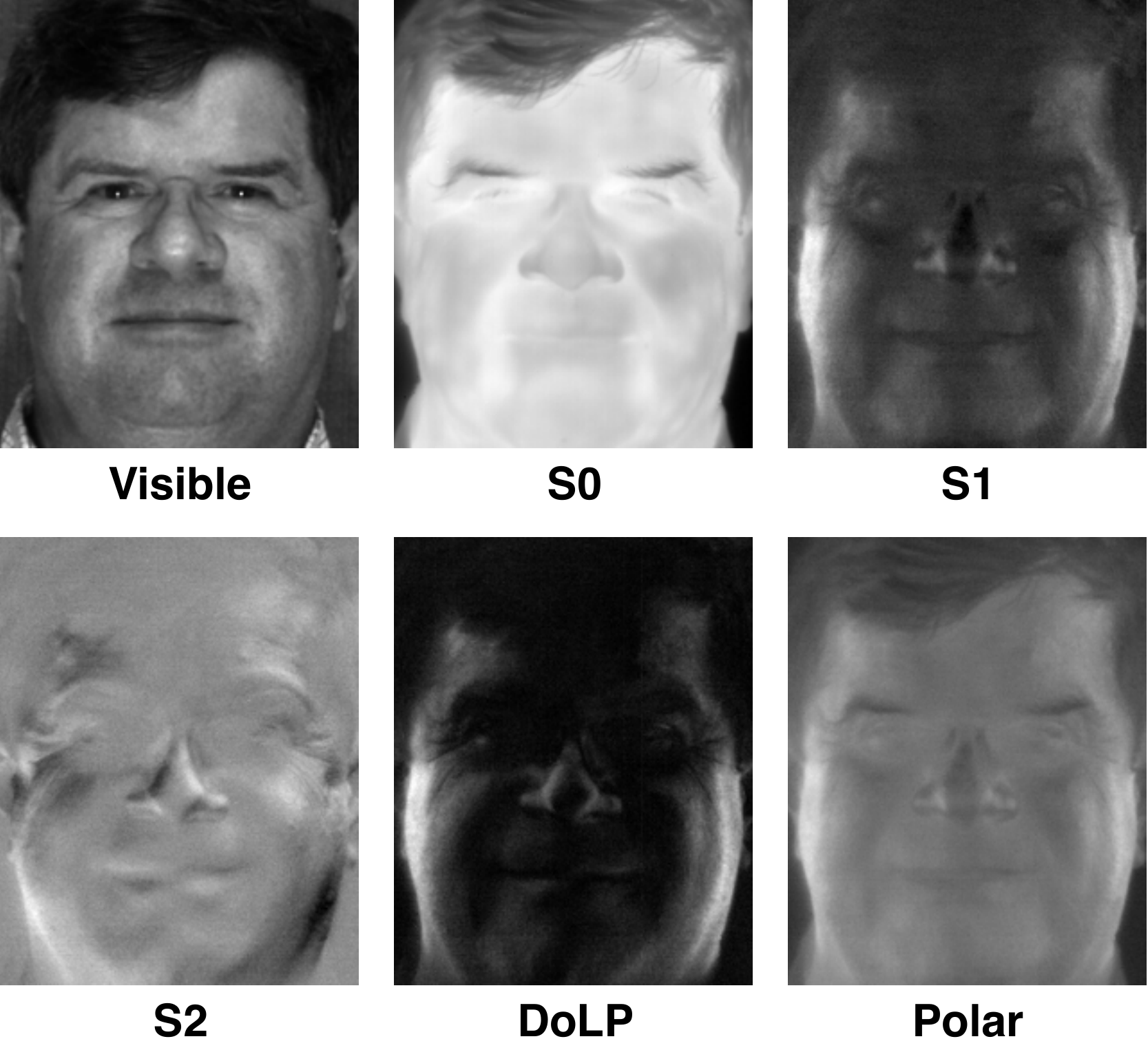}
	\caption{Sample Stokes as well as polarimetric and visible images corresponding to a subject in the ARL dataset \cite{hu2016polarimetric}. }
	\label{fig:datasample}
\end{figure}

Despite the success of CNN-based methods in addressing various challenges in FR, they are fundamentally limited to recognize
face images that are collected near-visible spectrum.  In many practical scenarios such
as surveillance in low-light conditions, one has to detect and recognize faces that are captured using thermal modalities  \cite{hu2016polarimetric,riggan2016estimation,short2015improving,zhang2017generative,Riggan2018thermal,klare2010heterogeneous,nicolo2012long, lezama2017not, bourlai2010cross}.  However, the performance of many deep learning-based methods degrades significantly when they are presented with thermal face images.  For example, it was shown in \cite{zhang2017generative,Riggan2018thermal} that simply using  deep features extracted from both raw polarimetric thermal and visible facial images are not sufficient enough for cross-domain face recognition.  The performance degradation is mainly due to the significant distributional change between the thermal and visible domains as well as a lack of sufficient data for training the deep networks for cross-modal matching. 

In many recent approaches, the polarization-state information of thermal emissions has been used to achieve improved cross-spectrum face recognition performance \cite{hu2016polarimetric,riggan2016estimation,short2015improving,zhang2017generative,Riggan2018thermal} since it captures geometric and textural details of faces that are not present in the conventional thermal facial images \cite{short2015improving,hu2016polarimetric}.    A polarimetric thermal image consists of four Stokes images: $S_{0}$, $S_{1}$, $S_{2}$, and degree-of-linear-polarization (DoLP), where $S_{0}$ indicates the conventional total intensity thermal image, $S_{1}$ captures the horizontal and vertical polarization-state information, $S_{2}$ captures the diagonal polarization-state information and DoLP describes the  portion  of  an  electromagnetic  wave  that  is  linearly polarized \cite{hu2016polarimetric}.   These Stokes images along with the visible and the polarimetric images corresponding to a subject in the ARL dataset \cite{hu2016polarimetric} are shown in Figure~\ref{fig:datasample}. It can be observed that $S_{1}$, $S_{2}$ and DoLP tend to preserve more textural details compared to $S_0$.  Similar to \cite{zhang2017generative, Riggan2018thermal}, we also refer to Polar as the three channel polarimetric image with $S_0$ , $S_1$ and $S_2$ as the three channels.   

Several attempts have been made to address the polarimetric thermal-visible face recognition problem \cite{Riggan2018thermal,riggan2016estimation,zhang2017generative}.  For instance, Riggan \etal \cite{riggan2016estimation} proposed a two-step procedure
(visible feature estimation and visible image reconstruction) to solve this cross-modal matching problem. Zhang \etal \cite{zhang2017generative} proposed an end-to-end generative adversarial network by fusing the different Stokes images as a multi-channel input to synthesize the visible image given the corresponding polarimetric signatures. Recently, Riggan \etal \cite{Riggan2018thermal} developed a global and local region-based technique to improve the discriminative quality of the synthesized visible imagery.  Though these methods are able to synthesize photo-realistic visible face images to some extent, the synthesized results in  \cite{zhang2017generative,reed2016generative,Riggan2018thermal} are still far from optimal and they tend to lose some semantic attribute information such as mouth open, mustache, etc.  Such reconstructions may degrade the performance of thermal to visible face verification.  

\begin{figure}
	\centering
	\includegraphics[width=0.95\linewidth]{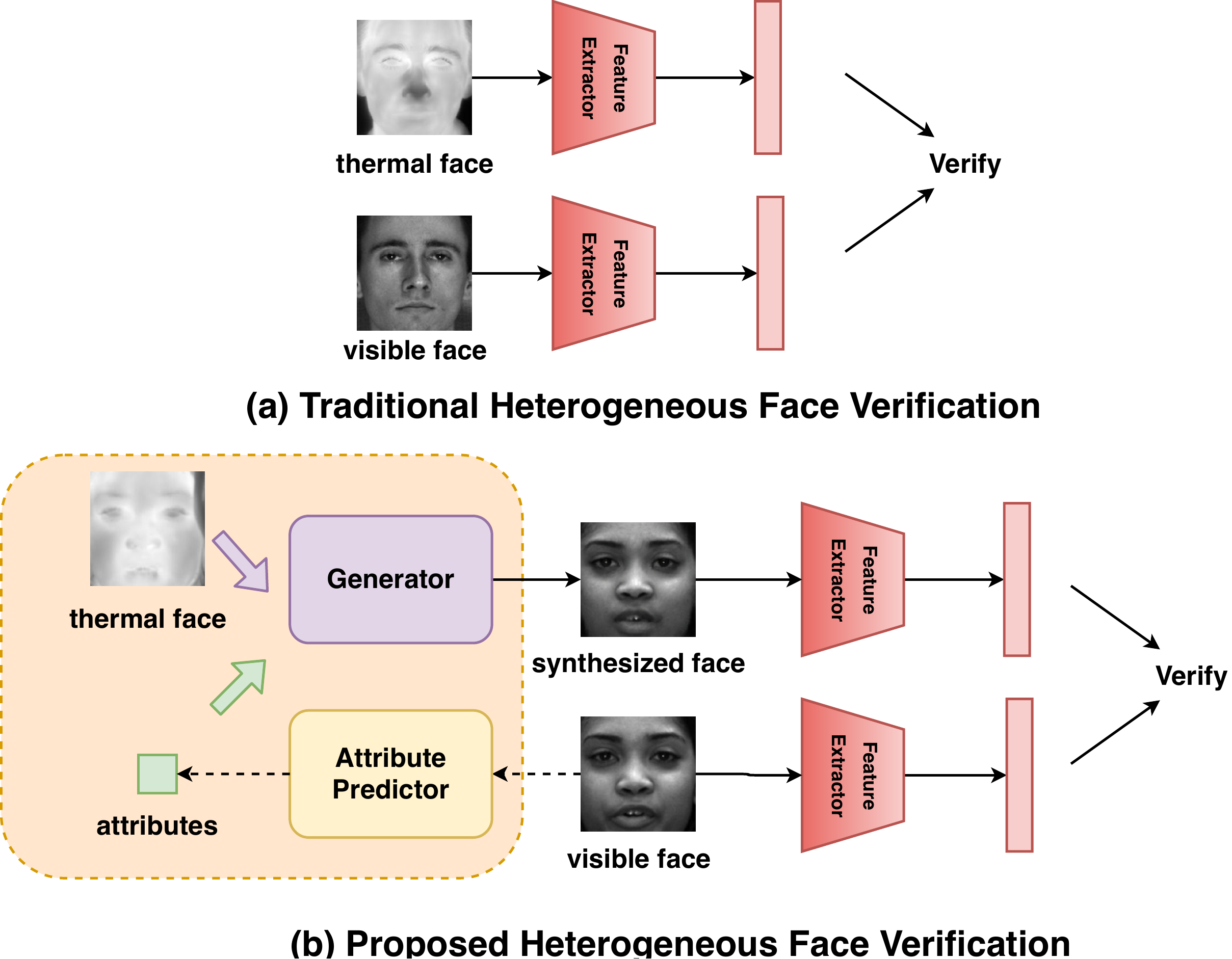}
	\caption{(a) Traditional heterogeneous face verification approaches use the features directly extracted from different modalities for verification \cite{hu2015thermal,klare2010heterogeneous,thermalfacerecognition2012,studymidwave2012}. (b) The proposed heterogeneous face verification approach uses a thermal face and semantic attributes to synthesize a visible face. Finally, deep features extracted from the synthesized and visible faces are used for verification.}
	\label{fig:framework}
\end{figure}

In this paper, we take a different approach to the problem of thermal to visible matching.  Figure~\ref{fig:framework} compares the traditional cross-modal verification problem with that of the proposed attribute-preserved cross-modal verification approach.  Given a visible and thermal pair, the traditional approach first extracts some features from these images and then verifies the identity based on the extracted features \cite{klare2010heterogeneous} (see Figure~\ref{fig:framework}(b)).  In contrast,  we propose a novel framework in which we make use of the attributes extracted from the visible image to synthesize the attribute-preserved visible image from the input thermal image for matching (see Figure~\ref{fig:framework}(b)).  In particular, a pre-trained VGG-Face model \cite{parkhi2015deep} is used to extract the attributes from the visible image.  Then, a novel Attribute Preserved Generative Adversarial Network (AP-GAN) is proposed to synthesize the visible image from the thermal image guided by the extracted attributes.  Finally, a deep network is used to extract features from the synthesized  and the input visible images for verification.  

\begin{figure*}
	\centering
	\includegraphics[width=0.95\linewidth]{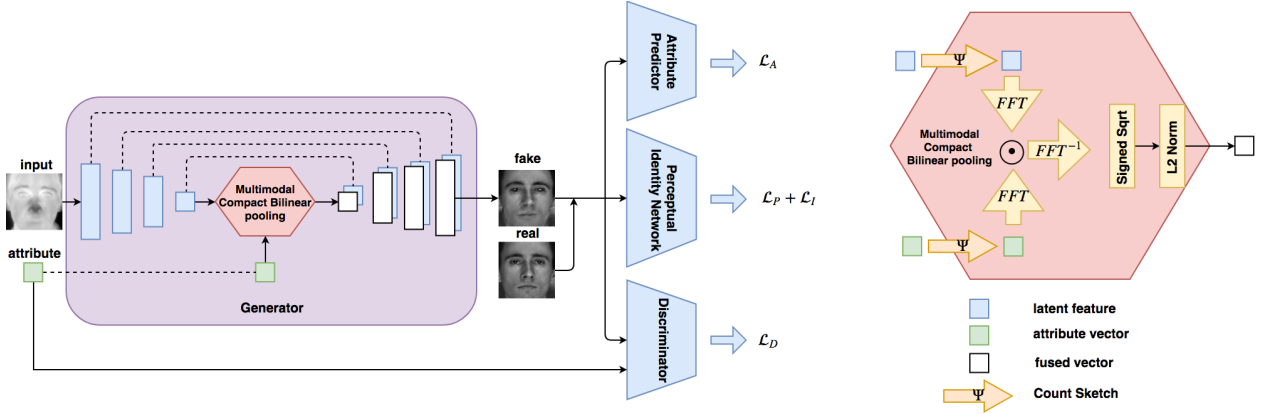}\\
	\raggedleft
	(a) AP-GAN Framework \hspace{40mm} (b) Multimodal Compact Bilinear (MCB) pooling
	\caption{ (a) A U-net based generator with MCB pooling is proposed to fuse the semantic attribute information with the image feature in the latent space. A triplet-pair is adopted for the discriminator in order to discriminate fake/real images as well as the corresponding semantic attributes. In order to generate high-quality and attribute-preserving images, a multi-purpose loss is optimized for training the network. (b) The architecture of MCB.  Here, $FFT$ indicates the Fast Fourier Transform and $FFT^{-1}$ indicates the inverse FFT.}
	\label{fig:framework3}
\end{figure*}

The proposed AP-GAN model is inspired by the recent image generation from attributes/text works \cite{reed2016generative,zhang2017generative,di2017face}.  The AP-GAN consists of two parts: (i) a multimodal compact bilinear (MCB) pooling-based generator \cite{fukui2016multimodal,gao2016compact}, and (ii) a triplet-pair discriminator.  The generator fuses the extracted attribute vector with the image feature vector in the latent space.  On the other hand, the discriminator uses triplet pairs (real image/true attributes, fake image/true attributes, real image/wrong attributes) to not only discriminate between real and fake images but also to discriminate between the image and the attributes.  In order to generate high-quality and attribute-preserved images, the generator is optimized by a multi-purpose objective function consisting of adversarial loss \cite{goodfellow2014generative},  $L_{1}$ loss, perceptual loss \cite{johnson2016perceptual},  identity loss \cite{zhang2017generative} and  attribute preserving loss. The entire AP-GAN framework is shown in Figure~\ref{fig:framework3}.

To summarize, the following are our main contributions:
\begin{itemize}

	\item  A novel thermal-visible face verification framework is proposed in which AP-GAN is developed for synthesizing visible faces from thermal (conventional or polarimetric)  images using facial attributes.
	
	\item A novel MCB pooling \cite{fukui2016multimodal,gao2016compact} based generator is proposed  to fuse the given attributes with the image features.

	\item A novel triplet-pair discriminator is proposed,  where  the discriminator \cite{reed2016generative} not only learns to discriminate between real/fake images but also to discriminate between the image and the corresponding semantic attributes.

	\item 	Extensive experiments are conducted on the ARL Facial Database \cite{hu2016polarimetric} and  comparisons
	are  performed  against  several  recent  state-of-the-art
	approaches. Furthermore, an ablation study is conducted to
	demonstrate the improvements obtained by including semantic attribute information for synthesis.

\end{itemize}

\section{Related Work} \label{related work}
In this section, we review some related works on thermal to visible face synthesis and recognition.  
 
\subsection{Traditional Thermal-Visible Face Recognition}

As described in Figure~\ref{fig:framework}, traditional thermal to visible face verification methods
first extract features from the visible and thermal images and then verify the identity based on the extRacted features.  Both hand-crafted and learned features have been investigated in the literature.   Hu \etal \cite{hu2015thermal} proposed a partial least squares (PLS) regression-based approach for cross-modal matching. Klare \etal \cite{klare2013heterogeneous} developed a generic framework for heterogeneous face recognition based on kernel prototype nonlinear similarities.   Another multiple texture descriptor fusion-based method was proposed by Bourlai \etal in \cite{studymidwave2012} for cross-modal face recognition. In \cite{thermalfacerecognition2012} PLS-based discriminant analysis approaches
were used to correlate the thermal face signatures to the visible face signatures. Some of the other visible to thermal cross-modal matching methods include  \cite{Gurton:14,short2015exploiting,Short:15}.

\subsection{Synthesis-based Thermal-Visible Face Verification}

Unlike the above mentioned traditional methods, synthesis-based thermal to visible face verification algorithms leverage the synthesized visible faces for verification.  Due to the success of CNNs and recently introduced generative adversarial networks (GANs) in synthesizing realistic images, various deep learning-based approaches have been proposed in the literature for thermal to visible face synthesis \cite{Riggan2018thermal,zhang2017generative,zhang2017tv,riggan2016estimation}. For example, Riggan \etal \cite{riggan2016estimation} proposed a two-step procedure
(visible feature estimation and visible image reconstruction) to solve the thermal-visible verification problem. Zhang \etal \cite{zhang2017generative} proposed an end-to-end GAN-based approach for synthesizing photo-realistic visible face images from their corresponding polarimetric images. Recently Riggan \etal \cite{Riggan2018thermal} proposed a new synthesis method to enhance the discriminative quality of generated visible face images by leveraging both global and local facial regions.

\section{Proposed Method} \label{proposed method}

\begin{figure*}[t]
\centering
\includegraphics[width=0.85\linewidth]{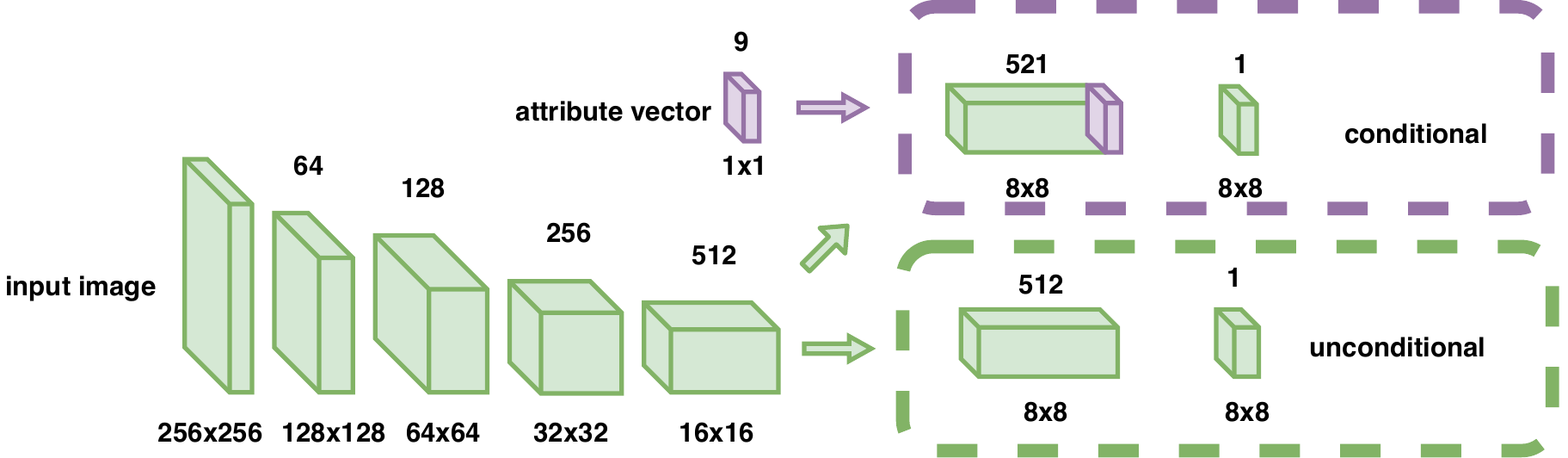}
\caption{An overview of the triplet-pair discriminator. The triplet-pair discriminator is composed of a conditional and an unconditional streams. The unconditional stream aims to discriminate the fake and real images. The conditional stream aims to discriminate between the image and the corresponding attributes.}
\label{fig:discriminator}
\end{figure*}

In this section, we discuss details of the proposed AP-GAN method. In particular, we discuss the proposed attribute predictor, generator and discriminator networks as well as the loss function used to train the network. 

\subsection{Attribute Predictor} \label{attribute predictor}
To efficiently extract attributes from a given visible face, an attribute predictor is fine-tuned based on the VGG-Face network \cite{parkhi2015deep} using the ten annotated attributes. This network is trained separately from AP-GAN. The fine-tuned network is used in both obtaining the visible face attributes and for capturing the attribute loss.

\subsection{Generator} \label{sec: generator}
A  U-net structure \cite{ronneberger2015u} is used as the building block for the generator since it is able to better capture large receptive field and also able to efficiently address the vanishing gradient problem.  In addition,  to effectively combine the extra facial attribute information into the building block, we fuse the attribute vector and the image feature in the latent space \cite{reed2016generative,zhang2017generative,di2017face}. Note that the attributes are extracted from the given visible face using the fine-tuned model as discussed above. The architecture corresponding to the generator is shown in Figure~\ref{fig:framework3}(a).

In our experiments, we observe that simple concatenation of the two vectors (encoded image vector and attribute vector) does not work well. One possible reason is that both vectors are significantly different in terms of their dimensionality.   Thus, we adopt the well-known MCB pooling method \cite{fukui2016multimodal,gao2016compact} to overcome this issue. Instead of simple concatenation, MCB leverages the following two techniques: bilinear pooling and sketch count. Bilinear pooling is the outer-product and linearization of two vectors, where all elements of both vectors are interacting with each other in a multiplicative way. In order to overcome the high-dimension computation of bilinear pooling, Pham \etal \cite{pham2013fast} implemented the count sketch of the outer product of two vectors, which involves the Fast Fourier Transform ($FFT$) and inverse Fast Fourier Transform ($FFT^{-1}$). The architecture of MCB module is shown in Figure~\ref{fig:framework3}(b). The generator network we use in this paper can be described as follows: \\
CL(64)-CBL(128)-CBL(256)-CBL(512)-CBL(512)-CBL(512)-CBL(512)-CBL(512)-MCB(512)-DBR(512)-DBR(512)-DBR(512)-DBR(512)-DBR(256)-DBR(128)-DBR(64)-DT(3),\\
where C stands for the convolutional layer (stride 2, kernel-size 4, and padding-size 1), L stands for Leaky Relu layer (negative\_slope=0.02), B stands for the batch-normalization layer,  MCB indicates the Multimodal Compact Bilinear module \cite{fukui2016multimodal,gao2016compact}, D stands for the deconvolutional layer (stride 2, kernel-size 4 and padding-size 1), R is the RuLU layer, and T is the Tanh function layer. All the numbers in parenthesis indicate the  channel number of the output feature maps.

\subsection{Discriminator} \label{sec: discriminator}
Motivated by the work \cite{pix2pix2017}, a patch-based discriminator $D$ is leveraged in the proposed method and it is trained iteratively with $G$. As discussed above, the discriminator $D$ not only aims to discriminate between real/fake images but also to discriminate between the image and the corresponding attributes.   Similar to the discriminator in  \cite{reed2016generative,StackGAN++}, a triplet pair is given to the discriminator: real-image/true-attributes (Real), synthesized-image/true-attributes (Fake), real-image/wrong-attributes (Fake). Given an input image $X$, and attribute vector $Y$, the overall objective function for training $D$ is as follows:
\begin{equation} \label{eq. discriminator loss}
\begin{split}
\mathcal{L}_{D_{uncond}} = -\frac{1}{2} E_{X\sim P_{data}(X)}[\log D(X)]+ \\ -\frac{1}{2} E_{G(X)\sim P_{G(X)}}[\log (1-D(G(X)))], \\
\mathcal{L}_{D_{cond}} = -\frac{1}{2} E_{X\sim P_{data}(X,A)}[\log D(X)]+ \\ -\frac{1}{2} E_{G(X)\sim P_{G(X)}}[\log (1-D(G(X),A))], \\
\mathcal{L}_{D} = \mathcal{L}_{D_{uncond}} + \mathcal{L}_{D_{cond}}, \qquad \qquad \qquad \quad
\end{split}
\end{equation}
where the unconditional loss $\mathcal{L}_{D_{uncond}}$ is to discriminate between real and synthesized samples. This information is back-propagated to $G$ to make sure the generated samples are as realistic as possible. In addition, the conditional loss $\mathcal{L}_{D_{cond}}$ is added to discriminate whether the given image matches the attributes. This information is back-propagated to $G$ so that it generates samples that are attribute preserving.

The architecture corresponding to the  discriminator is shown in Figure~\ref{fig:discriminator}. It consists of 6 convolutional blocks for both conditional and unconditional streams. Details of these convolutional blocks are as follows: \\
  NCL(64)-NCBL(128)-NCBL(256)-NCBL(512)-CBL(512)-CS(1),\\
where N stands for the Gaussian noise layer used to improve the training stability, with zero-mean and standard derivation of 0.01.  S stands for the sigmoid activation layer. Note that the only difference between the unconditional and conditional stream is the concatenation of the attribute vector at the fifth convolutional block.

\subsection{Object Function} \label{sec: stretagy}

The generator is optimized by minimizing the following loss
\begin{equation} \label{eq. all generator loss}
\mathcal{L}_{AP-GAN} = \mathcal{L}_{G} + \mathcal{L}_{A} + \lambda_{P}\mathcal{L}_{P} + \lambda_{I}\mathcal{L}_{I} + \lambda_{1}\mathcal{L}_{1},
\end{equation}
where $\mathcal{L}_{G}$ is the adversarial loss for generator $G$, $\mathcal{L}_{P}$ is the perceptual loss, $\mathcal{L}_{I}$ is the identity loss, $\mathcal{L}_{A}$ is the attribute loss,
$\mathcal{L}_{1}$ is the loss based on
the $L_{1}$-norm between the target and the reconstructed image, $\lambda_{P}, \lambda_{I}, \lambda_{A}, \lambda_{1}$ are weights respectively for perceptual loss, identity loss, attribute loss and $L_{1}$ loss. 

\subsubsection{Adversarial Loss}
Similar to the discriminator $D$, the adversarial loss for the  generator $G$ consists of both conditional and unconditional parts as defined below 

\begin{equation} \label{eq. generator loss}
\begin{split}
\mathcal{L}_{G} = \frac{1}{2} E_{G(X)\sim P_{G(X)}}[\log (1-D(G(X)))] + \\
\frac{1}{2} E_{G(X)\sim P_{G(X)}}[\log (1-D(G(X),A))].
\end{split}
\end{equation}
The generator $G$ therefore jointly approximates the image distribution conditioned (or unconditioned) on the attributes $A$.

\subsubsection{Perceptual and Identity Loss}
Perceptual loss was introduced by Johnson \etal \cite{johnson2016perceptual} for style transfer and super-resolution.  It has been observed that the perceptual loss produces visually pleasing results than $L_{1}$ or $L_{2}$ loss.  The perceptual and identity losses are defined as follows
\begin{equation}\label{perceptual, identity loss}
\mathcal{L}_{P,I} = \sum_{c=1}^{C}\sum_{w=1}^{W}\sum_{h=1}^{H}\|V(G(X))^{c,w,h}-V(Y)^{c,w,h}\|_{2},
\end{equation}
where $V$ represents a non-linear CNN feature. VGG-16 \cite{simonyan2014very} is used to extract features in this work. $C,W,H$ are the dimensions of features from a certain level of the VGG-16, which are different for perceptual and identity losses. 

In addition, $L_{1}$ loss between the synthesized image $G(X)$ and the real image $Y$ is used to capture the low-frequency information, which is defined as follows 
\begin{equation} \label{eq:L1 loss}
\mathcal{L}_{1} = \| G(X) - Y \|_{1}.
\end{equation} 

\begin{table}[h]
	\centering
	\caption{The facial attributes used in this work. }
	\begin{tabular}{|c|c|}
		\hline attributes & \makecell{Arched\_Eyebrows, Big\_Lips, Big\_Nose,\\
			Bushy\_Eyebrows, Male, Mustache,  \\ Narrow\_Eyes, No\_Beard, \\Mouth\_Slightly\_Open, Young}  \\ 
		\hline 
	\end{tabular} 
	\label{tb:attributes table}
\end{table}

\subsubsection{Attribute Loss}
Inspired by the perceptual loss, we define an attribute preserving loss, which measures the error between the attributes of the synthesized image and the real image. To make sure the pre-trained model captures the facial attribute information,   we  fine-tune the pretrained VGG-Face network on the attribute dataset and regard the fine-tuned attribute classifier as the pre-trained model for the  attribute preserving loss. Similar to the perceptual loss, the  $\mathcal{L}_A$ is defined as follows 
\begin{equation}\label{eq:attribute loss}
\mathcal{L}_{A} = \|Q(G(X))-Q(Y)\|_{2},
\end{equation}
 where $Q$ is the fine-tuned attribute predictor network and $N$ is the total number of output neurons. By feeding such an attribute information into the generator during training, the generator $G$ is able to learn semantic information corresponding to the face. 

\begin{table*}[t]
	\caption{Verification performance comparisons among the baseline methods, state-of-the-art methods, and the proposed AP-GAN method for both polarimetric thermal (Polar) and conventional thermal (S0) cases.}
	\centering
	\begin{tabular}{|c|c|c|c|c|}
		\hline Method & AUC(Polar) & AUC(S0) & EER(Polar) & EER(S0) \\ 
		\hline Raw & $50.35\%$ & $58.64\%$ & $48.96\%$ & $43.96\%$ \\ 
		\hline Mahendran \etal \cite{mahendran2015understanding} & $58.38\%$ & $59.25\%$ & $44.56\%$ & $43.56\%$ \\ 
		\hline Riggan \etal \cite{riggan2016estimation}  & $75.83\%$  & $68.52\%$ & $33.20\%$ & $34.36\%$ \\ 
		\hline Zhang \etal \cite{zhang2017generative} & $79.90\%$  & $79.30\%$  & $25.17\%$  & $27.34\%$ \\ 
		\hline Riggan \etal \cite{Riggan2018thermal} & $85.43\%$ & $82.49\%$ & $21.46\%$ & $26.25\%$ \\ 
		\hline AP-GAN (\textbf{our}) & $\mathbf{88.93\%\pm 1.54\%}$ & $\mathbf{84.16\%\pm 1.54\%}$ & $\mathbf{19.02\%\pm 1.69\%}$ & $\mathbf{23.90\%\pm 1.52\%}$\\ 
		\hline AP-GAN (\textbf{GT}) & $\mathbf{91.28\%\pm 1.68\%}$ & $\mathbf{86.08\%\pm 2.68\%}$ & $\mathbf{17.58\%\pm 2.36\%}$ & $\mathbf{23.13\%\pm 3.02\%}$ \\ 
		\hline 
	\end{tabular}
	\label{tb: comparison_table}
\end{table*}

\subsection{Implementation} \label{sec: implementation}

The entire network is trained in Pytorch on a single Nvidia Titan-X GPU. During the AP-GAN training, the $L_{1}$, perceptual and identity loss parameters are chosen as  $\lambda_{1}=10$, $\lambda_{P}=2.5$, $\lambda_{I}=0.5$, respectively. The ADAM \cite{kingma2014adam} is implemented as the optimization algorithm with parameter $betas=(0.5, 0.999)$ and batch size is chosen as 3. The total epochs are 200. For the first 100 epochs, we fix the learning rate as $0.0002$ and for the remaining 100 epochs, the learning rate was decreased by $1/100$ after each epoch. The feature maps for the perceptual and the identity loss are from the relu1-1 and the relu2-2 layers, respectively.  In order to fine-tune the attribute predictor network, we manually annotate images with the attributes tabulated in Table~\ref{tb:attributes table}.

\section{Experimental Results} \label{experimental result}
The proposed method is evaluated on the ARL Multimodal Face Database \cite{hu2016polarimetric} which consists of polarimetric  (i.e. Stokes image) and visible images from 60 subjects.  Similar to the protocol discussed in \cite{riggan2016estimation}, we only use the images from Range 1 and their corresponding attributes are obtained from fine-tuned attribute predictor network.  In particular, Range 1 images from 30 subjects and the corresponding attributes are used for training. The remaining 30 subjects' data are used for evaluation. We repeat this process 5 times and report the average results.

We evaluate the face verification performance of proposed method compared with several recent works \cite{zhang2017generative,Riggan2018thermal,pix2pix2017}. Moreover, the performance is evaluated on the FC-7 layer of the  pre-trained VGG-Face model \cite{parkhi2015deep} using the receiver operating characteristic (ROC) curve, Area Under the Curve (AUC) and Equal Error Rate (EER) measures. To summarize, the proposed method is evaluated on the following two protocols:\\
\indent (a) Conventional thermal (S0) to Visible (Vis).\\
\indent (b) Polarimetric thermal (Polar) to Visible (Vis).

\begin{figure*}[t]
	\centering
	\begin{minipage}{.48\textwidth}
		\centering
		\includegraphics[width=0.9\linewidth]{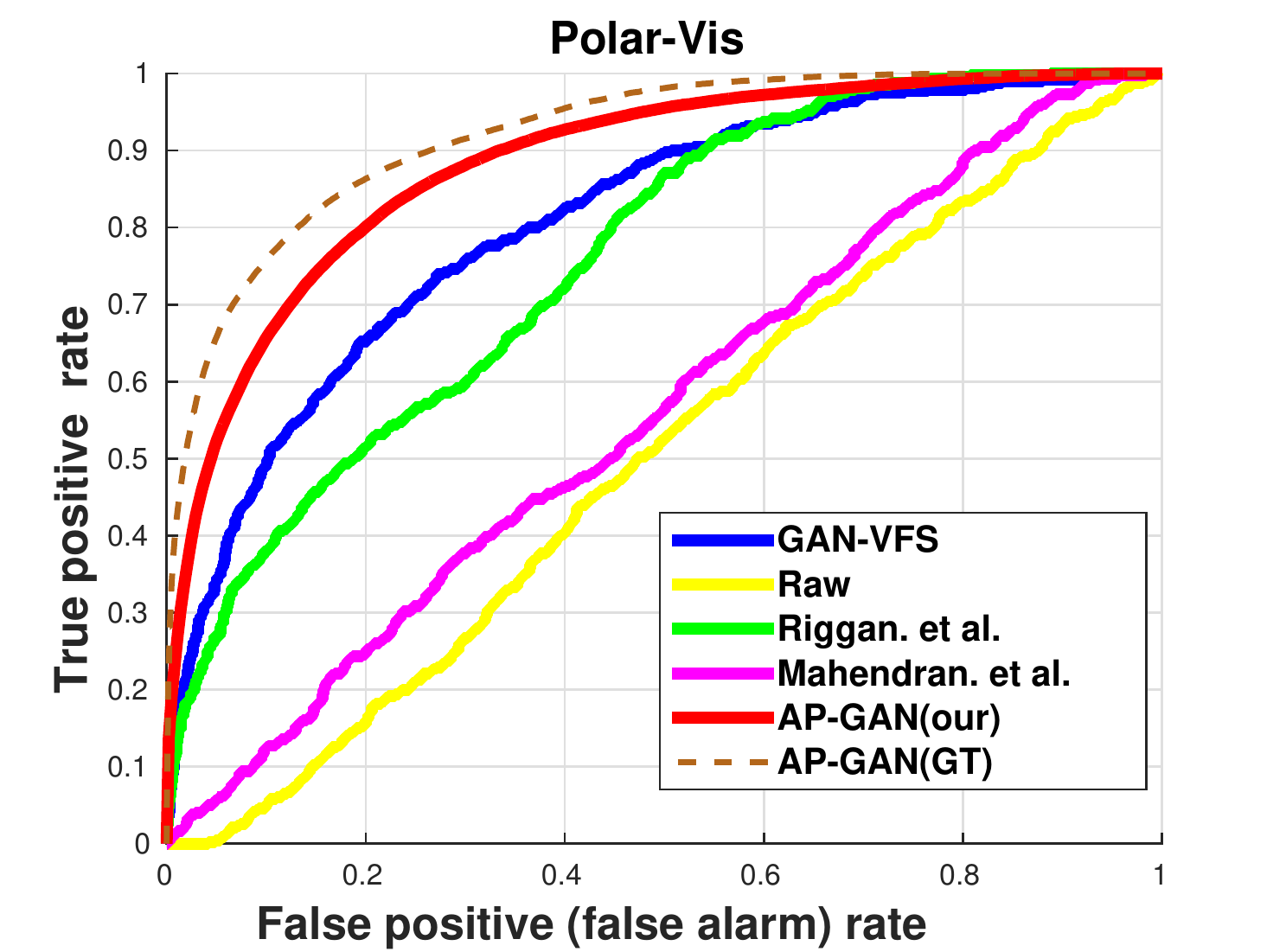}\\
		(a)
	\end{minipage}
	\begin{minipage}{.48\textwidth}
		\centering
		\includegraphics[width=0.9\linewidth]{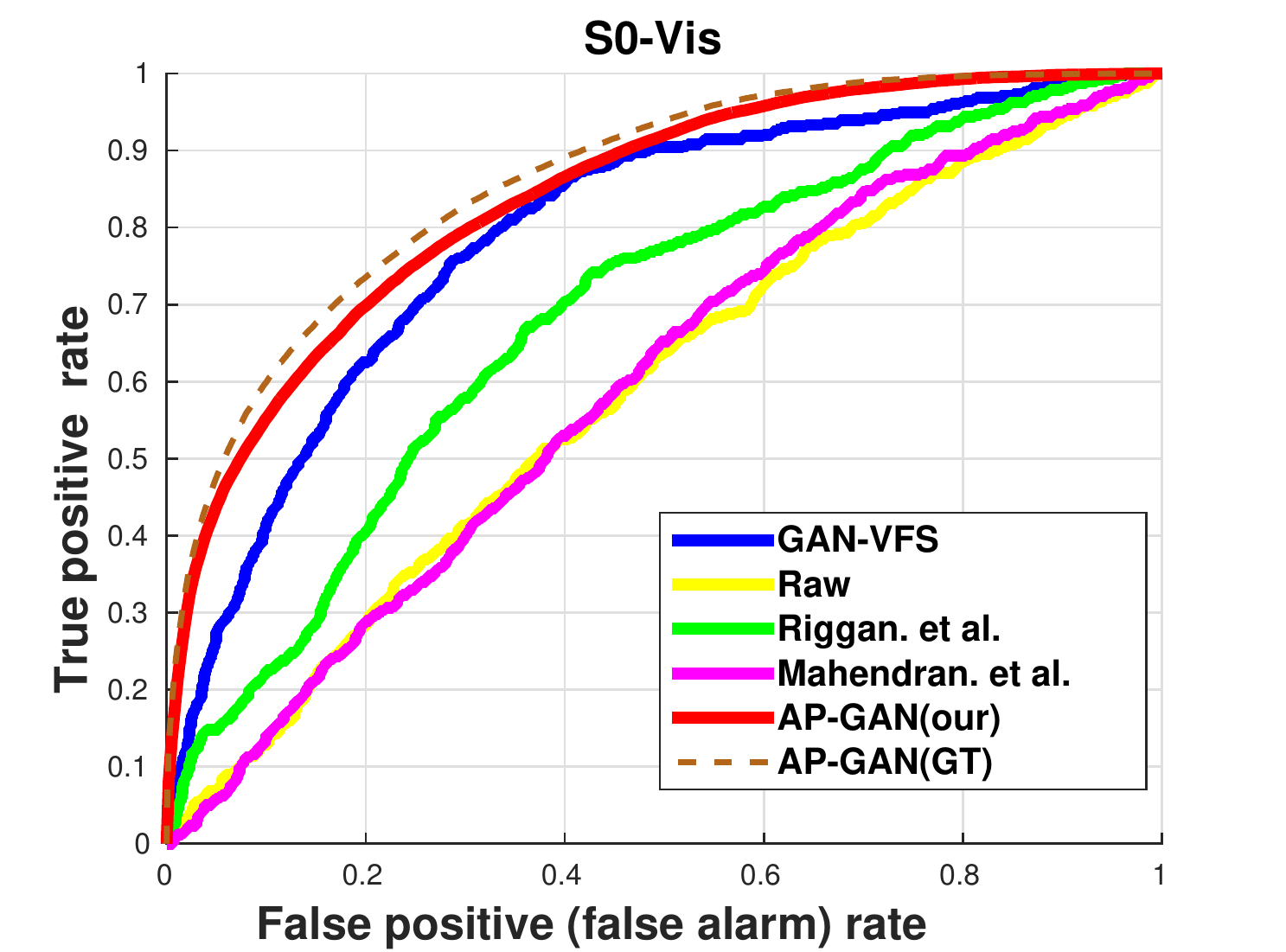}\\
		(b)
	\end{minipage}
	\caption{The ROC curve comparison with several state-of-the-art methods: (a) The performance on Polar-Visible verification. (b) The performance  on S0-Visible verification. }
	\label{fig:Comparison_Figure}
\end{figure*}

\begin{figure*}[t]
	\centering
	\includegraphics[width=0.95\linewidth]{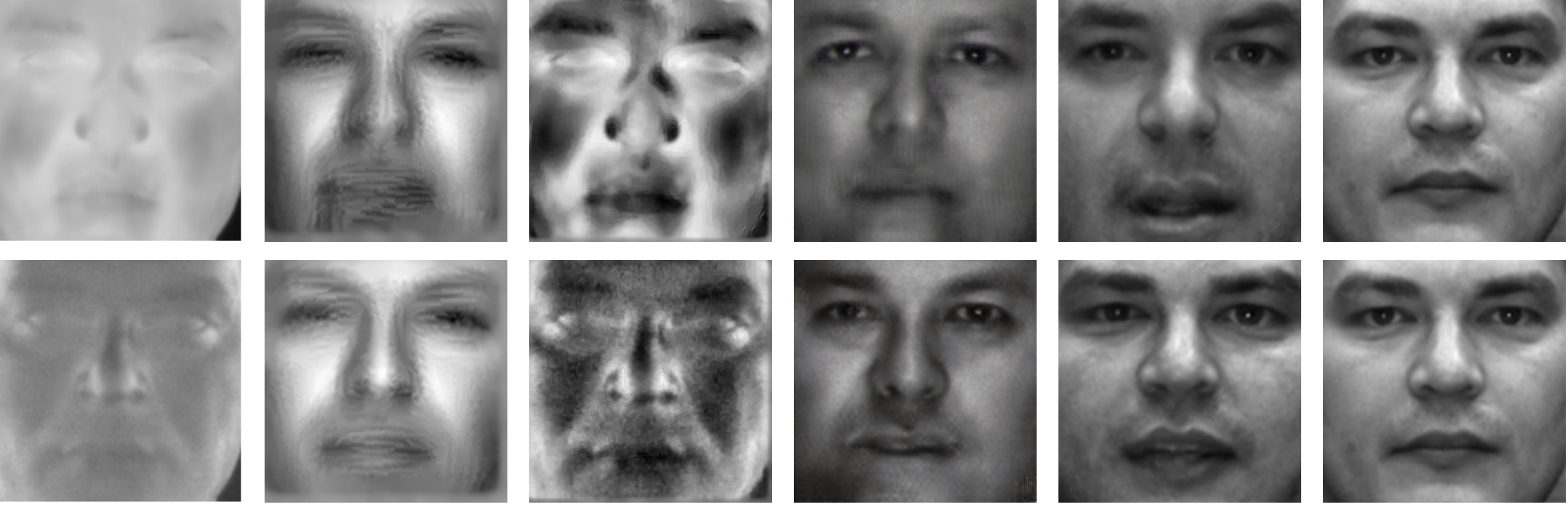}\\
	\raggedright
	\small
	\hspace{13mm} Input \hspace{12mm} Riggan \etal \cite{riggan2016estimation} \hspace{2mm} Mahendran \etal \cite{mahendran2015understanding} \hspace{2mm} Zhang \etal \cite{zhang2017generative}\hspace{6mm} \textbf{AP-GAN (ours)} \hspace{6mm} Ground Truth\\
	\caption{The synthesized samples from different methods: Riggan \etal \cite{riggan2016estimation}, Mahendran \etal \cite{mahendran2015understanding}, Zhang \etal \cite{zhang2017generative},  \textbf{AP-GAN (ours)}, ground truth. The first row results correspond to the S0 image, and the second row results correspond to the Polar image.}
	\label{fig:comparison_figure}
\end{figure*}

\subsection{Preprocessing}
In addition to the standard preprocessing in  \cite{hu2016polarimetric}, two more pre-processing steps are used for the proposed method. First, the faces in visible images are detected by MTCNN \cite{mtcnn}. Then, a standard central crop method is used to crop the detected faces. Since the MTCNN is implementable on the visible images only, we use the same detected rectangle coordinations to crop the S0, S1, S2 images. After preprocessing, all the images are scaled to be $256\times256$ and saved as 16-bit PNG files.

\begin{figure*}[t]
	\centering
	\includegraphics[width=0.9\linewidth]{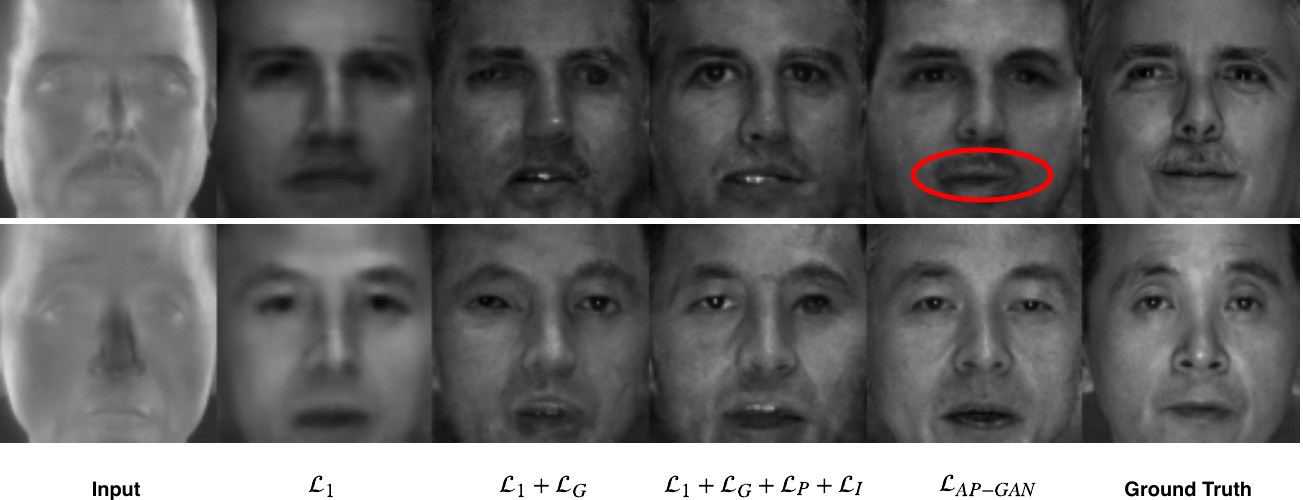}
	\caption{The visual results of the ablation study for different experimental settings. Given input Polar image, synthesized results using different combination of losses are shown successively from left to right columns as $\mathcal{L}_{1}$; $\mathcal{L}_{1}+\mathcal{L}_{G}$;  $\mathcal{L}_{1}+\mathcal{L}_{G} +\mathcal{L}_{P}+\mathcal{L}_{I}$; $\mathcal{L}_{AP-GAN}$. The ground truth images are on the most right side.}
	\label{fig:ablation_figure}
\end{figure*}

\subsection{Comparison with state-of-the-art Methods}
We evaluate and compare the performance of the proposed method with that of recent  state-of-the-art methods \cite{zhang2017generative,mahendran2015understanding,riggan2016estimation,Riggan2018thermal}. In addition to our method, we also conduct experiments with a baseline method '\textbf{AP-GAN(GT)}' where we use the ground truth attributes in our method rather than automatically predicting them using the proposed attribute predictor. This baseline will clearly determine how effective the proposed attribute predictor is in determining the attributes from unconstrained visible faces.

Figure~\ref{fig:Comparison_Figure} shows the evaluation performance for two different experimental settings, S0 and Polar separately. Compared with other state-of-the-art methods in Figure~\ref{fig:Comparison_Figure}, the proposed  method performs better with a larger AUC and lower EER scores. In addition, it can be observed that the performance corresponding to the Polar modality is better than the S0 modality, which also demonstrates the advantage of using the polarimetric thermal images than the conventional thermal images. The quantitative comparisons, as shown in the Table~\ref{tb: comparison_table}, also demonstrate the effectiveness of proposed method. 


In addition to the quantitative results, we also show some visual comparisons in Figure~\ref{fig:comparison_figure}. The first row in Figure~\ref{fig:comparison_figure} shows one synthesized  sample using S0. The second row shows the same synthesized  sample using Polar. It can be observed that results of  Riggan \etal \cite{riggan2016estimation} do capture the overall face structure but it tends to lose some details on the skin. Results of Mahendran \etal \cite{mahendran2015understanding} poor compared to \cite{riggan2016estimation}.  Results of Zhang \etal \cite{zhang2017generative} are more photo-realistic but tend to lose some attribute information. The proposed AP-GAN not only generates photo-realistic images but also preserves attributes on the reconstructed images.

\subsection{Ablation Study}

\begin{figure}[t]
	\centering
	\includegraphics[width=0.85\linewidth]{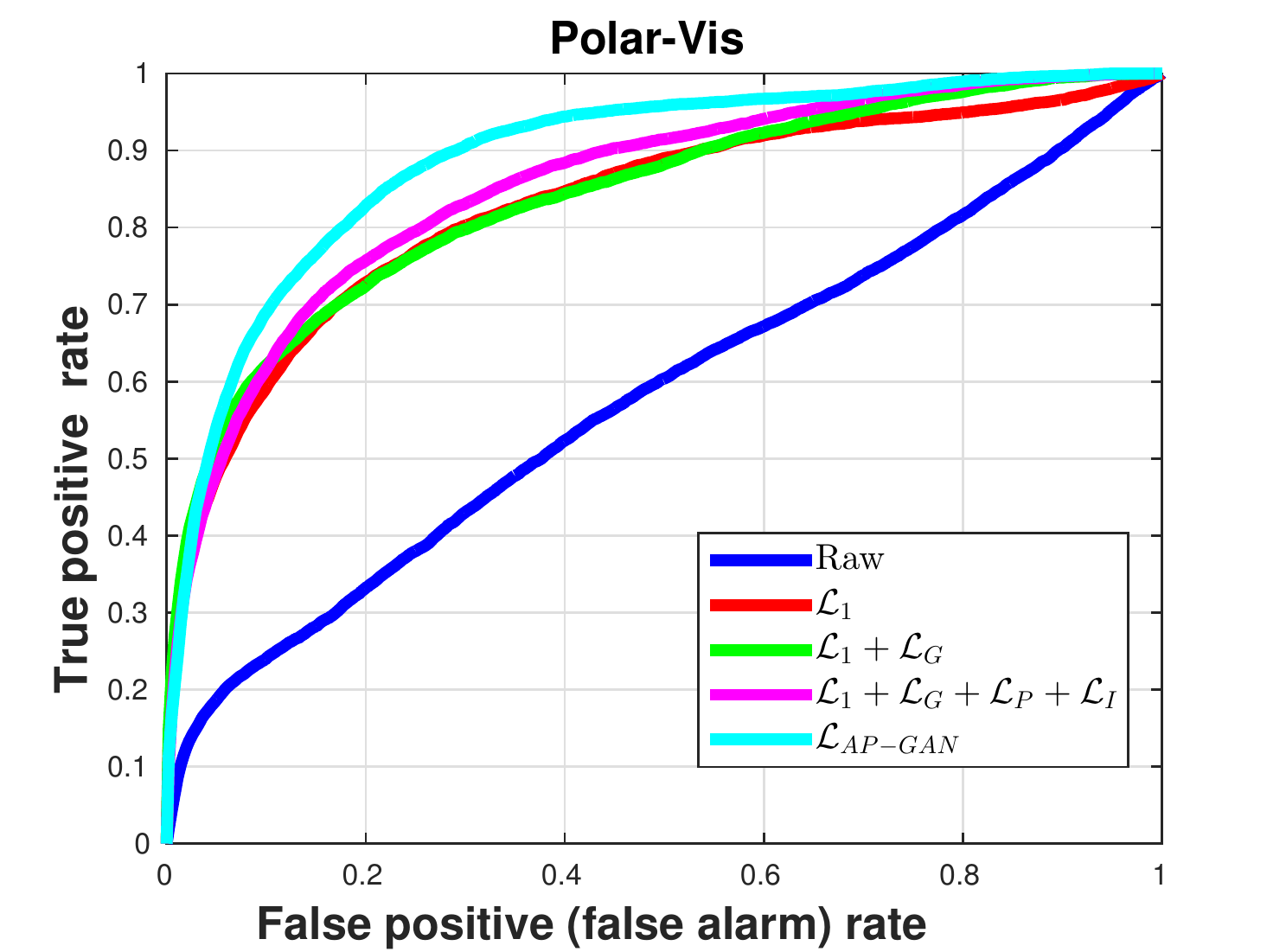}
	\caption{The ROC curves corresponding to the ablation study.}
	\label{fig:Ablation_Polar}
\end{figure}

In order to demonstrate the effectiveness of different modules in the proposed method, we conduct the following ablation studies: (1) Polar to Visible estimation with only $\mathcal{L}_{1}$ loss, (2) Polar to Visible estimation with $L_{1}$ and adversarial loss $\mathcal{L}_{G}$, (3) Polar to Visible estimation with $\mathcal{L}_{1}$, $\mathcal{L}_{G}$, and perceptual and identity loss  $\mathcal{L}_{P},\mathcal{L}_{I}$, (4) Polar to Visible estimation with all the losses $\mathcal{L}_{AP-GAN}$ as defined in  Eq.~\eqref{eq. all generator loss}. Figure~\ref{fig:Ablation_Polar} shows the ROC curves corresponding to each experimental setting. All the experiments in the ablation study are evaluated from one experimental split of the Polar modality. From this figure, we can observe that  using all the losses together as $\mathcal{L}_{AP-GAN}$, we obtain the best performance. Compared to the results of $\mathcal{L}_{AP-GAN}$ and $\mathcal{L}_{1}+\mathcal{L}_{G} + \mathcal{L}_{P} + \mathcal{L}_{I}$, we can clearly see the improvements obtained by fusing the semantic attribute information with the image feature in the latent space.

Besides the ROC curves, we also show the visual results for each experimental setting in Figure~\ref{fig:ablation_figure}. Given the input Polar image, the synthesized results from different experimental setting are shown in Figure~\ref{fig:ablation_figure}. It can be observed that  $\mathcal{L}_{1}$ captures the low-frequency features of images very well. $\mathcal{L}_{1} + \mathcal{L}_{G}$ can capture both low-frequency and high-frequency features in the image. However, it adversely introduced distortions and artifacts in the synthesized image. In addition, optimizing $\mathcal{L}_{P} + \mathcal{L}_{I}$ suppresses these distortions to some extent. Finally,  fusing attributes into the previous loss can not only improving the performance but also preserves facial attributes, like the mustache as shown in the red circle.

\subsection{Attribute Manipulation Result}

Instead of visually and quantitatively showing the performance of AP-GAN on face verification, we also show results when the attributes are manipulated.

\begin{figure}
	\centering
	\includegraphics[width=0.95\linewidth]{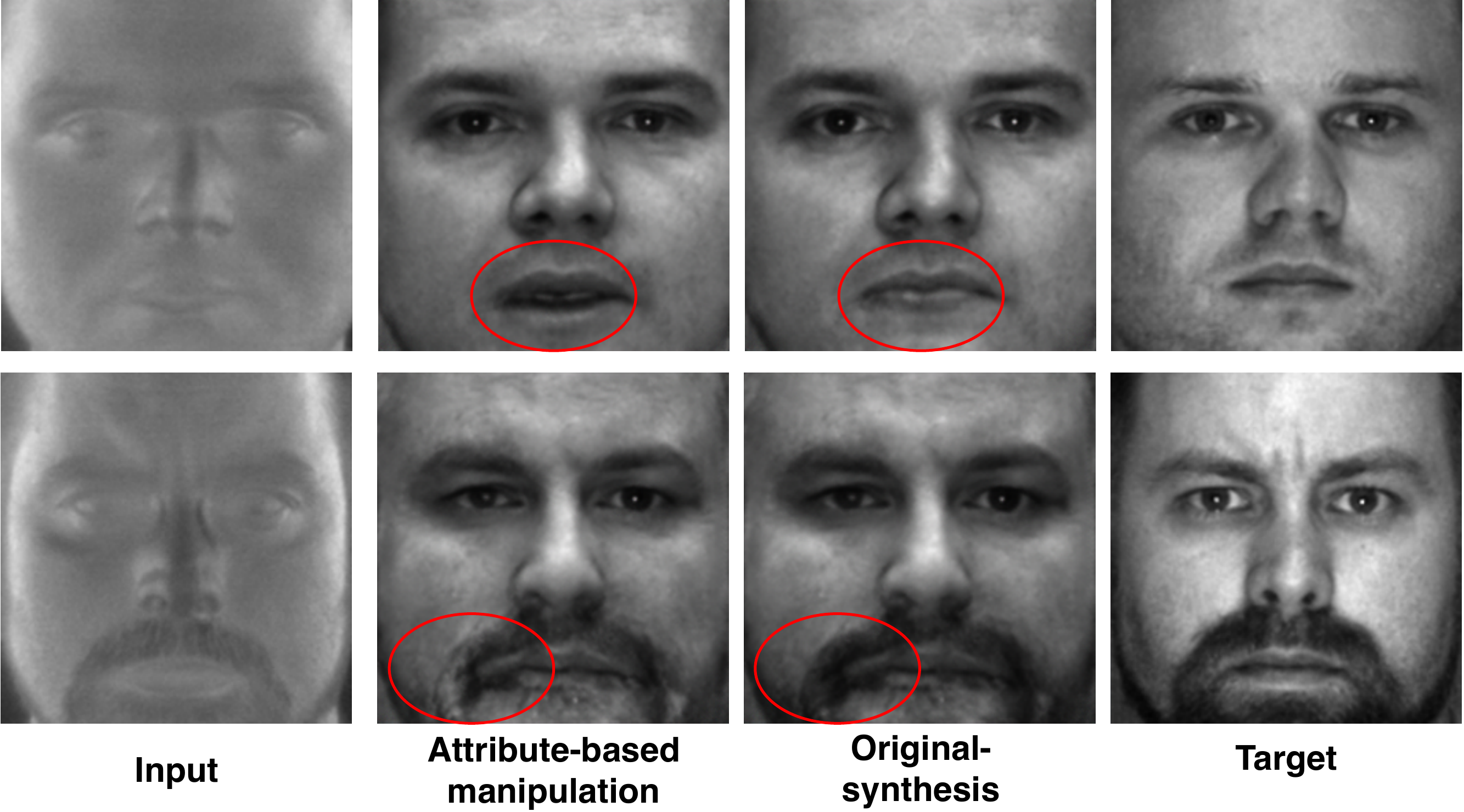}
	\caption{Attribute manipulation results. Given the input thermal image and manipulated attributes, the synthesized visible image shows the changes caused by these attributes. The first row shows the visual difference on mouth\_open attribute changing from $-1$ to $1$ (the other attributes are fixed). The second row shows the visual difference on mustache attribute changing from $1$ to $-1$ (the other attributes are fixed). The attribute-based manipulation and original-synthesis column show the comparison with and without attribute change, respectively.}
	\label{fig:manipulate_result}
\end{figure}

Given a certain thermal image, by manipulating its corresponding attributes, we obtain some interesting synthesis results as shown in Figure~\ref{fig:manipulate_result}. In the first row of Figure~\ref{fig:manipulate_result}, the \emph{mouth\_open} attribute value was changed from $-1$ to $1$ while the other attribute values were fixed.  As can be seen from the generated figure,  the synthesized image shows a slightly open mouth. In the second row, we show the resutls for changing the attribute value corresponding to mustache from $1$ to $-1$. The generated results clearly capture the attribute change as shown with a red circle.

\section{Conclusion} \label{conclusion}

We propose a novel Attribute Preserving Generative Adversarial Network (AP-GAN) structure for polarimetric-visible face verification via synthesizing photo realistic visible face images from the corresponding thermal (polarimetric or conventional) images with extracted attributes. Rather than use only image-level information for synthesis and verification, we take a different approach in which semantic facial attribute information is also fused during training and testing. Quantitative and visual experiments evaluated on a real thermal-visible dataset demonstrate that the proposed method achieves state-of-the-art performance compared with other existing methods. In addition, an ablation study is developed to demonstrate the improvements obtained by different combination of loss functions.

\section*{Acknowledgement}
This work was supported by an ARO grant W911NF-16-
1-0126.

{\small
\bibliographystyle{ieee}
\bibliography{btas2018_copy}
}

\end{document}